\documentclass[11pt]{article}
\usepackage{authblk}
\usepackage{coling2020}
\usepackage{times}
\usepackage{url}
\usepackage{graphicx}
\usepackage{latexsym}
\usepackage{multicol}
\usepackage{multirow}
\usepackage{amsmath}
\usepackage{subfig}
\usepackage{wrapfig}
\usepackage{babel,blindtext}
\usepackage{booktabs}
\usepackage{rotating}
\usepackage{tabularx}
\usepackage[dvipsnames]{xcolor}

\usepackage{hyperref}
\definecolor{ultramarine}{rgb}{0.07, 0.04, 0.56}

\hypersetup{
    colorlinks=true,
    linkcolor=ultramarine,
    citecolor=ultramarine,
    filecolor=ultramarine,  
    urlcolor=ultramarine
}

\newcolumntype{Y}{>{\centering\arraybackslash}X}

\newcolumntype{M}[1]{>{\centering\arraybackslash}m{#1}}

\usepackage{titlesec}
\usepackage[titletoc,toc,title]{appendix}

\newcommand{\sx}[0]{\spadesuit}

\newcommand{\sz}[0]{\diamondsuit}
\newcommand{\sw}[0]{\heartsuit}
\def\code#1{\texttt{#1}}

\colingfinalcopy 

\title{Deconstruct to Reconstruct a Configurable Evaluation Metric for Open-Domain Dialogue Systems}

\author[$\sx $]{\textbf{Vitou Phy}}
\author[$\sz$]{\textbf{Yang Zhao}}
\author[$\sw \sx$]{\textbf{Akiko Aizawa}}
\affil[$\sx $]{The University of Tokyo, Japan}
\affil[$\sz $]{IBM-Research Tokyo, Japan}
\affil[$\sw $]{National Institute of Informatics, Japan}
\affil[ ]{\texttt{\{vitou, aizawa\}@nii.ac.jp}, 
\texttt{yangzhao@ibm.com}}

\date{}

\begin{document}
\maketitle

\begin{abstract}

Many automatic evaluation metrics have been proposed to score the overall quality of a response in open-domain dialogue. Generally, the overall quality is comprised of various aspects, such as relevancy, specificity, and empathy, and the importance of each aspect differs according to the task. For instance, specificity is mandatory in a food-ordering dialogue task, whereas fluency is preferred in a language-teaching dialogue system. However, existing metrics are not designed to cope with such flexibility. For example, BLEU score fundamentally relies only on word overlapping, whereas BERTScore relies on semantic similarity between reference and candidate response. Thus, they are not guaranteed to capture the required aspects, i.e., specificity. To design a metric that is flexible to a task, we first propose making these qualities manageable by grouping them into three groups: understandability, sensibleness, and likability, where likability is a combination of qualities that are essential for a task. We also propose a simple method to composite metrics of each aspect to obtain a single metric called \textbf{USL-H}, which stands for \textbf{U}nderstandability, \textbf{S}ensibleness, and \textbf{L}ikability in \textbf{H}ierarchy\footnote{The implementation of our metrics is available at \url{https://github.com/vitouphy/usl_dialogue_metric}.}. We demonstrated that USL-H score achieves good correlations with human judgment and maintains its configurability towards different aspects and metrics.

\end{abstract}

\section{Introduction}

Evaluating a dialogue response is crucial for the development of open-domain dialogue systems. It allows comparison between different systems, which is similar to how the machine translation community uses BLEU \cite{papineni2002bleu} to evaluate the overall quality of the translation and determines whether a system is the state-of-the-art \cite{Bahdanau2015NeuralMT,sennrich2016neural,aharoni2019massively}. Without automatic evaluation metrics, many studies tend to rely on either expert or crowdsourced platform to score the responses, which are both time-consuming and cost-ineffective \cite{zhang2018learning,zhan2019ssa,adiwardana2020towards}. To cope with this, various metrics have been proposed to score the overall quality of a dialogue response. 

Word overlap-based metrics, which were adopted from the MT community to measure the overlapping words between reference and candidate sentences, have been used to evaluate the dialogue responses \cite{sordoni2015neural,zhang2018learning}. However, \newcite{liu2016not} showed that these metrics, i.e., BLEU \cite{papineni2002bleu}, METEOR \cite{banerjee2005meteor}, or ROUGE score \cite{Lin:2004}, do not correlate well with human judgements, because there are many possible responses to reply to a given context. Recently, learning-based metrics, which aim to predict the overall quality of a response, have a better correlation with human judgment compared with word overlap-based metrics. Various training settings have also been explored. For example, ADEM \cite{lowe2017towards} is trained to predict the score by learning to regress on human judgments, whereas PONE \cite{lan2020pone} is trained with next utterance prediction task with sophisticated samplings.

However, these metrics are not configurable and may suffer from several limitations. First, they may not capture a certain quality that is essential for a particular task. As shown in Table \ref{table:example_not_capture}, BERTScore and BERT-RUBER tend to assign a relatively high score to the unspecific response. This might be due to the complexity of the overall score. Generally, a single overall score is usually comprised of different qualities, such as readability, specificity, and empathy, and the importance of each aspect differs according to the task. For example, specificity is preferred in food-ordering chatbots, whereas fluency is preferred in language-teaching chatbots. However, the existing metrics are not flexible to such changes. BERTScore \cite{Zhang2020BERTScore:}, for example, relies on using pre-trained BERT embedding \cite{devlin2019bert} to compute similarity between reference and candidate responses; thus this does not guarantee good correlation for the specificity quality (Table \ref{table:example_not_capture}). Another limitation is the difficulty in enhancing only a particular aspect of the metric. Suppose there is a single metric that can capture both sensibleness and specificity, and a new state-of-the-art metric on the latter quality is subsequently developed; it would be complicated to modify the existing metrics (i.e., BLEU or ADEM) to include this new SOTA metric. 

Aside from evaluating a response using only a single overall score, some studies evaluate the response on multiple aspects, i.e., fluency, relevancy, specificity, and empathy \cite{zhang2018learning,weston2018retrieve,smith2020can}. The limitation of this approach is that with multiple scores to consider, it becomes unclear to determine which response is better. Is a specific response more preferable than an empathetic one? 

\begin{table}[t]
\centering

\begin{minipage}[t!]{0.55\linewidth}
\small
\centering
    \begin{tabular}{p{0.9\textwidth}}
    \toprule
    \textbf{Context:} \\
    I'm sorry. It's out of stock now. Could you come by again next week?
    \\\midrule
    \textbf{Responses:} \\
    
    \begin{tabularx}{0.95\linewidth}{rl}
    Ground-truth: & Next week? It is too late. I need it urgently. \\
    Candidate 1: & Yes. That would be fine. \textcolor{ultramarine}{[Unspecific]}\\
    & \textbf{B}: 0.12, \textbf{R}: 0.97, \textbf{U}: 0.67, \textbf{H}: 0.75 \\ 
    Candidate 2: & Sure. What day is best to come by? \textcolor{ultramarine}{[Specific]}\\
    & \textbf{B}: 0.02, \textbf{R}: 0.98, \textbf{U}: 0.71, \textbf{H}: 1.0 \\
    \end{tabularx} \\
    \bottomrule
   \end{tabular}
    \caption{Examples on how metrics on overall quality may not capture specificity. B, R, U, and H denotes scores from BERTScore, BERT-RUBER, USL-H (proposed), and human, respectively.}
    \label{table:example_not_capture}
\end{minipage}\hfill
\begin{minipage}[t!]{0.40\linewidth}
\centering
\includegraphics[width=\linewidth]{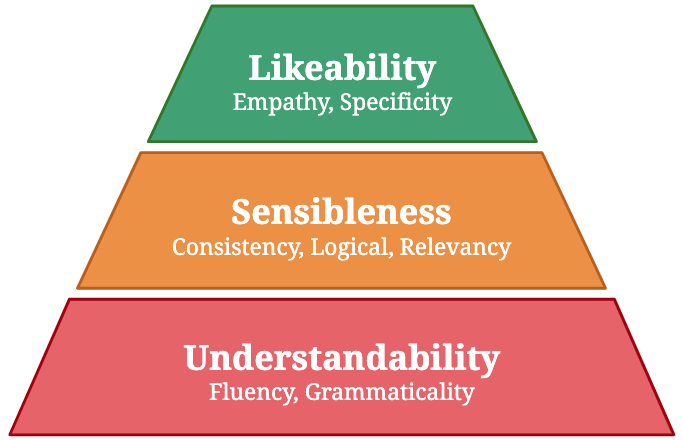}
\captionof{figure}{Decomposition the structure of a response quality.}
\label{fig:hiearchy}
\end{minipage}
\end{table}

To address these issues, we first propose simplifying the various qualities by grouping them into three main aspects: understandability \cite{nubel1997end}, sensibleness \cite{adiwardana2020towards}, and likability. We assume these groups have hierarchical properties in the following way: (i) a response first needs to be \emph{understandable}; (ii) then it needs to \emph{make sense} to the context; (iii) other qualities, i.e., specificity, are just additional qualities that make an acceptable response more \emph{likable} for a given task (Figure \ref{fig:hiearchy}). If we want the score to capture empathy instead, we only need to replace specificity in the top layer with empathy. In other words, the likability aspect does not need to implicitly capture the understandability or sensibleness, as it will be checked by the lower layers in the hierarchy. Based on these properties, we propose a simple method to combine scores of each aspect to obtain \textbf{USL-H} score, which stands for \textbf{U}nderstandability, \textbf{S}ensibleness, and \textbf{L}ikability in \textbf{H}ierarchy. 
USL-H can be modified to add or remove a quality and to replace a metric with a more optimal alternative. 
This configurability removes the barrier of requiring a single complicated model and instead enables a combination of multiple sub-metrics. 

For simplicity, we demonstrate the configurability using only specificity as our likability aspect. Experimenting on the DailyDialog dataset \cite{li2017dailydialog}, we show that valid utterance prediction, next utterance prediction, and masked language models have good correlations with human judgments on understandability, sensibleness, and specificity, respectively. Moreover, combining these sub-metrics as a single metric using our USL-H also correlates well with overall quality on both Pearson and Spearman correlations. Alternatively, we replace specificity with the empathy aspect, recombine the sub-metrics, and put it up against other metrics to select the most empathetic responses. We find that this new configuration can detect better empathetic responses compared to the rest. Through various experiments, we show that USL-H is configurable to capture certain qualities of a response and can be improved further upon replacing a sub-metric with a better performing alternative.

The main contributions of this paper are the following: (i) the grouping of various qualities of dialogue responses into three main aspects: understandability, sensibleness, and likability, (ii) introducing a configurable hierarchical evaluation metric that can be modified to work with a set of response's qualities and sub-metrics according to the task while achieving good correlation with human judgments.

\section{Related Work}

\paragraph{Automatic Evaluation Metrics} Many automatic evaluation metrics have been proposed to evaluate the overall quality of a response. BLEU \cite{papineni2002bleu}, METEOR \cite{banerjee2005meteor}, and ROUGE \cite{Lin:2004} metrics are word overlap-based approaches, which utilize the model's responses and reference responses to compute the number of overlapping words. The more words overlap, the higher the scores are. However, \newcite{liu2016not} showed that these metrics have a weak correlation with human judgment. Alternatively, embedding-based metrics \cite{wieting2016towards,rus2012comparison,forgues2014bootstrapping,Zhang2020BERTScore:} are used as measurements in the previous studies, for which the embeddings of context and response are used to obtain the similarity score \cite{zhang2018learning,zeng2019dirichlet}. However, due to many possible responses to a context, it is inaccurate to use these metrics.

Learning-based metrics have been explored recently, especially with the next utterance prediction setting \cite{tao2018ruber,ghazarian2019better,lan2020pone}. The model learns to determine whether or not a context-response pair is valid. It is typically trained with context-response pairs that appear in the dialogue as the positive examples. Then, negative sampling is used to obtain negative examples. Training using this setting demonstrates a moderate correlation with human judgment. However, since such learning-based metrics rely on the positive and negative examples, a sophisticated sampling technique is required to obtain appropriate examples that reflect a certain quality.

\paragraph{Score Composition} Some studies have attempted to develop metrics by combining scores from different aspects into a single score \cite{zhan2019ssa,adiwardana2020towards}. \newcite{zhan2019ssa} proposed a metric that combines scores from semantic and syntactic sub-metrics with a weighted average. This metric had a moderate correlation with human judgment and outperformed all of the word-overlap-based metrics. However, it does not consider qualities, such as specificity or diversity.

Alternatively, \newcite{adiwardana2020towards} proposed a human evaluation metric that considers both sensibleness and specificity, where specificity is dependent on sensibleness and is scored only if the response is sensible; otherwise, it is zero by default. Then, they obtained the final human overall score by averaging them together. Unlike \newcite{zhan2019ssa}, they did not use any metric for any aspect. Instead, they suggest using perplexity as the evaluation metric to capture both qualities. However, using a single metric, like perplexity, to monitor multiple aspects is not configurable when we want to evaluate another aspect, for instance, sensibleness and empathy. 


\section{Evaluation Criteria}
\label{sec:criteria}

\subsection {Fundamental Aspects}
The overall quality of a response contains various qualities, such as readability, fluency, relevancy, sensibleness, and specificity. However, not every aspect is equally important. A response may contain an interesting detail, but such information is not usable if it is completely off-topic. Likewise, if a response is not understandable, we suspect that it is difficult to determine whether it is a suitable reply. Based on this observation, we propose to cluster the qualities into three groups --- understandability \cite{nubel1997end}, sensibleness \cite{adiwardana2020towards}, and likability --- as illustrated in Figure \ref{fig:hiearchy}. We assume these aspects to be independent from each other, except for sensibleness, which we will discuss later.

\paragraph{Understandability} quantifies whether or not a response is understandable. A response does not have to be grammatically correct to be considered understandable due to the nature of conversations being compact, unstructured, and noisy (i.e., ``Not really''). Also, it does not need to be interesting, nor does it need context to be understandable. Due to such independency, we consider this aspect as the fundamental building block among the three groups. 


\paragraph{Sensibleness} measures how suitable a response is to a given context \cite{adiwardana2020towards}. For example, the response to the context ``Dinner's ready!'' can be short (``10 minutes''), generic (``Okay''), or intriguing (``It smells good already''). Any of these responses are considered as sensible. This quality is comprised of relevancy, consistency, common sense, and more. However, in this work, we do not focus on these sub-qualities. Instead, we consider the sensibleness quality as a whole. Please note that although a sensible response can be generic, short, and boring, it is rather important for the response to be on-topic than to have unique tokens. Thus, we place this quality on the second level of the hierarchy. 



\paragraph{Likability} quantifies how much a set of one or more qualities makes a response more likable for a particular task. These qualities can be diversity \cite{li2016diversity}, sentiment \cite{rashkin2019towards}, specificity \cite{ke2018generating}, engagement \cite{yi2019towards}, fluency \cite{kann2018sentence} and more. A likable response may or may not be sensible to the context. For example, a diverse response may contain many unique words, although it might be off-topic or completely incomprehensible.  However, when combining with sensibleness, it can quantify how likable a sensible response is. Due to the enhancement that the likability aspect has on the response, we position it on the highest level of the hierarchy.

\subsection{USL-H Metric}
\label{sec:usl_metric}

Each aspect, by itself, is not enough to evaluate the overall quality of a response. An understandable response might not be relevant to the context. A sensible response might be bland and generic. A response with many unique words (diversity) or rare words (specificity) does not guarantee that it is understandable or sensible. On the contrary, incorporating these qualities together with the following concept gives more useful information. First, we examine if the response is understandable. Then, we check whether it makes sense to the context. After that, we determine how likable that response is, i.e., how many unique or rare words in the response. If the response fails at any aspect of the hierarchy, the subsequent aspect will not be considered. With such construction, the likability score does not need to capture understandability nor sensibleness. Those criteria will be checked by aspects in the lower hierarchy. Such composition allows flexibility and configurability in utilizing different metrics, as it is not needed to search for a single metric that satisfies multiple aspects. Instead, we find metrics for those multiple aspects and combine them together with our proposed hierarchy to get a single metric. 



Formally, let us denote $s_U$, $s_S$, $s_L$ for scores of understandability, sensibleness, and likability, respectively, and $s_L$ can be comprised of one or more qualities $q_j$, i.e., specificity or empathy. In prior work, to reconstruct scores together, \newcite{zhan2019ssa} uses a weighted average to combine syntactic and semantic scores, whereas \newcite{adiwardana2020towards} uses the arithmetic average to combine the sensibleness and specificity scores. Particularly, \newcite{adiwardana2020towards} considers specificity being dependent on sensibleness such that if sensibleness is 0, so is specificity. Although they limit only to sensibleness and specificity, we extend these simple heuristics with our hierarchy concept into the following equation:


\begin{equation}
s_{USL\text{-}H} = \alpha_1 s_U + \alpha_2 s_U s_S + \alpha_3 s_U s_S s_L
\label{eq:fal}
\end{equation}

\begin{equation}
s_{L} = \sum \beta_j q_j
\label{eq:weight_sum}
\end{equation}
where $s_U$, $s_S$, $s_L$, $q_j$ are continuous variables ranging between $[0,1]$. $\sum \alpha_i = \sum \beta_j = 1$. $\alpha_i$ and $\beta_j$ are coefficients for each quality. These formulations can be applied to obtain scores for both automatic and human evaluations. 

There are two intuitions behind this heuristic. (i) Understandability score $s_U$ adds in clarity and interpretability when the response is unsensible. Otherwise, it is difficult to determine whether the unsensibleness is due to the response being completely incomprehensible or off-topic. (ii) Likability brings in other qualities that are not covered by the other two aspects, although it is considered in the final score only if the response is understandable and sensible. There is one key property of likability. Suppose the response already makes sense to the context. In that case, the likability score does not have to be context-dependent, as it is just an extra quality on top of a response that is already sensible. This composition allows flexibility in using different metrics, i.e., context-independent or context-dependent, for this likability aspect. To sum up, we consider understandability and sensibleness being context-independent and context-dependent, respectively, whereas likability being either of them. 




\paragraph{Simplified USL-H} During our preliminary analysis, we found that having $s_U$ in all the three terms is unstable. If the automatic evaluation metric corresponds to understandability misevaluates the response, the other qualities will be disregard. To alleviate such instability, we make an assumption that if a response is not understandable, it is unlikely that it is sensible ($s_S \approx s_U s_S$). In other words, sensibleness, at the observation level, is dependent  on understandability. Thus, we do not need $s_U s_S$, as $s_U$ is already captured by $s_S$ implicitly. With this assumption, we arrive at equation \ref{eq:fal2}, which we will use for the remaining of the paper. 




\begin{equation}
s_{USL\text{-}H} = \alpha_1 s_U + \alpha_2 s_S + \alpha_3 s_S s_L
\label{eq:fal2}
\end{equation}

\section{Automatic Evaluation Metrics} \label{sec:models}

\subsection{Problem Setting}
Each dialogue $D$ is comprised of {$u_1, u_2, \dots, u_n$} utterances, where each utterance contains $(w_1, w_2, \dots, w_m)$ words. Two consecutive utterances, $u_i$ and $u_{i+1}$, where $i < n$, are selected to form a context-response pair $(c,r_0)$, with $c$ as the context and $r_0$ as the ground-truth response. For each context $c$, we use a different generative or retrieval system, as described in Section \ref{sec:experiment}, to obtain a candidate response $r$.

\subsection{Baseline Metrics for Overall Quality}

\paragraph{Word-Overlap-based Metrics} We use \code{BLEU} \cite{papineni2002bleu}, \code{ROUGE-L} \cite{Lin:2004}, and \code{METEOR} \cite{banerjee2005meteor} to measure the word-overlapping score between $r_0$ and $r$.

\paragraph{Embedding-based Metrics} Different responses may contain different lexical words, although they may share a similar meaning. Thus, we also experiment with embedding-based metrics by comparing semantic information between $r_0$ and $r$ using the following metrics: \code{Embedding Averaging} \cite{wieting2016towards}, \code{Greedy Matching} \cite{rus2012comparison}, \code{Vector Extrema} \cite{forgues2014bootstrapping}, and \code{BERTScore} \cite{Zhang2020BERTScore:}.

\paragraph{Learning-based Metrics}
We also include reference-free automatic evaluation metrics, which have recently emerged as a topic of interest. We will use (i) \code{BERT-RUBER} \cite{ghazarian2019better}, which computes the embeddings for $c$ and $r$ and predicts the probability of whether these two are the right pair; (ii) \code{PONE} \cite{lan2020pone}, an extension of the \code{BERT-RUBER} metric, which utilizes generative responses as augmentation for positive labels and BERT-based retrieval responses as negative responses. However, through our experiments, we were not able to reproduce the performance mentioned in the study. Thus, we only use \code{PONE} with augmented negative responses, denoted as \code{EN-PONE}.

\subsection{Metrics for Fundamental Aspects}

\paragraph{Metric for Understandability} 
We train a model using a valid utterance prediction (\code{BERT-VUP}) setting to capture the understandability of an utterance $u$ by classifying whether or not it is valid. Unlike a sentence, which should be grammatically correct, an utterance does not need to satisfy this property, and the auxiliary verb or punctuation may be missing. We use these properties to build a training set of valid and invalid utterances. First, we randomly determine if $u$ should be valid. If it is, we will assign that with label one and randomly apply one of the following rules: (i) remove punctuation at the end, (ii) remove stop words, or (iii) no modification. Alternatively, we label it as zero and apply one of the following rules from \newcite{sinha2020learning} to obtain a negative sample: (i) word reorder (shuffle the order of all words), (ii) word drop (randomly drop x\% words), or (iii) words repeat (randomly select span(s) of words and randomly repeat them up to 3 times). For an utterance $u$ with $(w_1, w_2, \dots, w_m)$ words, we fine-tune BERT \cite{devlin2019bert} by obtaining the contextual embedding $h_i$ for each word $w_i$ and using max-pooling to obtain the utterance-level embedding. Then, we use a softmax layer to obtain the probability and use it as the final score $s_U$. 

\paragraph{Metric for Sensibleness} We train another model using the next utterance prediction (\code{BERT-NUP}) task as the metric for the sensibleness. Given a context-response pair $(c,r)$, the objective of the model is to classify whether that pair is a valid pair or not. To build label data for this binary classification task, we uses two consecutive utterances $(u_i, u_{i+1})$ from a dialogue $D$, where $u_i$ is the context $c$ and $u_{i+1}$ is its corresponding response $r$, and label them as a valid pair. Then, we keep $u_i$ as the context and select a random utterance $u_j$ from a pool of all the utterances in the training set and label that pair $(u_i, u_j)$ as the invalid pair. To fine-tune the BERT model, we first merge a context-response pair into a single array of tokens $(w_1, w_2, \dots, w_t)$. Then, we use the same approach as \code{BERT-VUP} metric to obtain the score $s_S$.

\paragraph {Metric for Specificity} For simplicity of studying the configurability of our proposed metric, we select specificity as our likable quality. Following the use of Roberta in \newcite{mehri2020usr} to compute the mask language model (\code{MLM}) metric, we use a BERT-based model for consistency with the \code{BERT-VUP} and \code{BERT-NUP} metrics. Moreover, instead of using both $(c, r)$, as in \newcite{mehri2020usr}, we only use the response $r$ to ensure the independence from the context $c$. Therefore, for a response $r$ with $m$ words, we sequentially mask one word at a time and feed it into \code{BERT-MLM} to predict negative log-likelihood (\code{MLM-Likelihood}) of all masked words. We also investigate negative cross-entropy (\code{MLM-NCE}), perplexity (\code{MLM-PPL}), and \code{MLM-SLOR} \cite{kann2018sentence} to verify if they can be used for the understandability and specificity aspects.

\section{Experiment}
\label{sec:experiment}

\paragraph{Training Corpus} The corpus used in this study is DailyDialog \cite{li2017dailydialog}, which is about day-to-day communication on everyday topics. This dataset consists of 11,118/1,000/1,000 dialogues for train/valid/test sets with explicit textual information of five dialogue acts and seven emotion labels. We split this dataset evenly into two parts: (i) for training generative and retrieval models to generate candidate responses, and (ii) for training automatic evaluation metrics for scoring each aspect.

\begin{table*}[t!]
\centering
\begin{tabular}{l|cccc} \toprule
& Understandable & Sensible & Specific & Overall \\ \midrule
Kappa & 0.4333 & 0.6110 & 0.4572 & 0.4137\\ \bottomrule
\end{tabular}
\caption{\label{font-table} Inter-annotator agreement on Cohen's Kappa. }
\label{inter-annotator-table}
\end{table*}

\paragraph{Building Response Candidates} To effectively evaluate the evaluation metrics, it is important to have a mix of good and bad responses for the metrics to score. Therefore, we choose two retrieval methods, two generative methods, and one human-generation for a total of five responses per given context. This includes TF-IDF, DualEncoder \cite{lowe2017training}, Seq2Seq with Attention Mechanism \cite{Bahdanau2015NeuralMT}, and DialoGPT \cite{zhang2019dialogpt}. These five responses vary in quality, i.e., generative models may produce incomprehensible or unspecific responses, whereas retrieval models may select unsensible responses.  Overall, we collected five responses from different models for 50 contexts, which accounts for 250 context-response pairs.

\paragraph{Human Judgement} It is necessary to evaluate if the evaluation metrics are comparable to human judgment. To verify this, we recruited four volunteers to collect the human judgment on the 50 contexts. For each context, five different responses from different models described in the previous section were presented for evaluation. The annotators were asked to score each context response pair with the following questions: (i) Is this response understandable \{0, 1\}?, (ii) Does this make sense to the context \{0, 1\}?, (iii) Does it at least have some detail \{0, 1\}?, (iv) Overall, how good is this response \{0,1,2,3\}? 

We also instructed the volunteers to consider these questions independently, with understandability and specificity independent from the context. Regarding evaluating the overall score, we did not provide fine-grained instructions of what each value represents. Instead, we only mentioned that the bad and good responses are corresponding with score of 0 and 3, respectively. How they score the responses is entirely subjective to each annotator. This allows us to observe how one would think if they were to judge the overall quality of a response. Then, we use Cohen's Kappa \cite{cohen1960coefficient} to measure pairwise inter-annotator agreements for all the aspects, presented in Table \ref{inter-annotator-table}. The annotators moderately agree on all qualities, with the lowest agreement on the overall score. This result is expected because no detailed instruction was provided to assist their annotations.

\paragraph{Experimental Setup} We use a pre-trained base-model of BERT to fine-tune for the BERT-VUP, BERT-NUP, and BERT-MLM metrics separately, by using the HuggingFace framework \footnote{\url{https://huggingface.co/}} on an NVIDIA Tesla V100 PCIe 32GB. These three models are trained with an ADAM optimizer \cite{KingmaB14} with a learning rate of 1e-5. We select the best version of each model based on the lowest validation loss.

\section{Results}

\begin{figure}[t!]

\begin{minipage}[t]{0.48\textwidth}
\includegraphics[height=4cm]{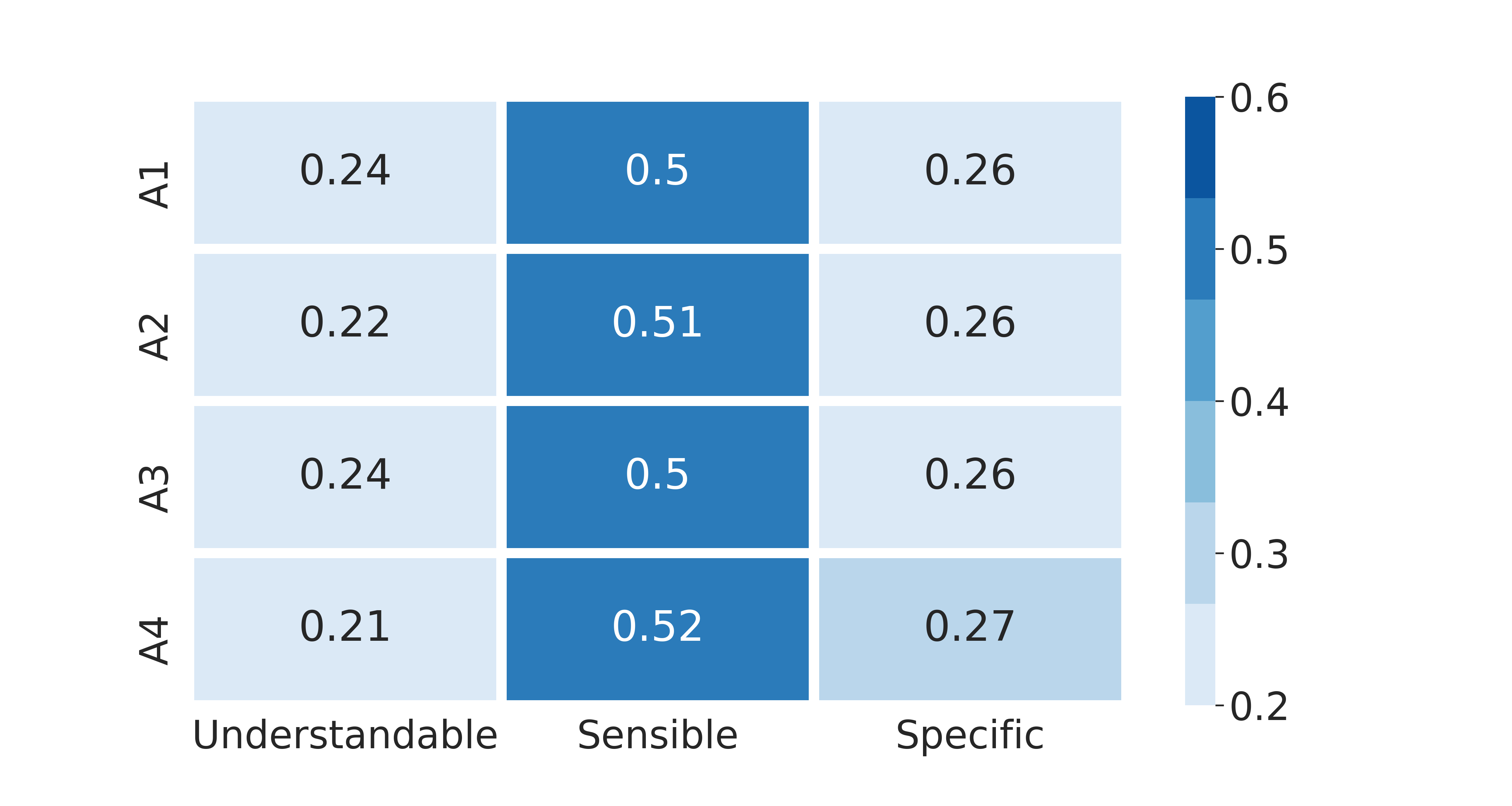}
\centering
\caption{Coefficient of each quality on the overall score per annotator $A_i$.}
\label{fig:annot_heatmap}
\end{minipage}\hfill
\begin{minipage}[t]{0.48\textwidth}
\includegraphics[height=4cm]{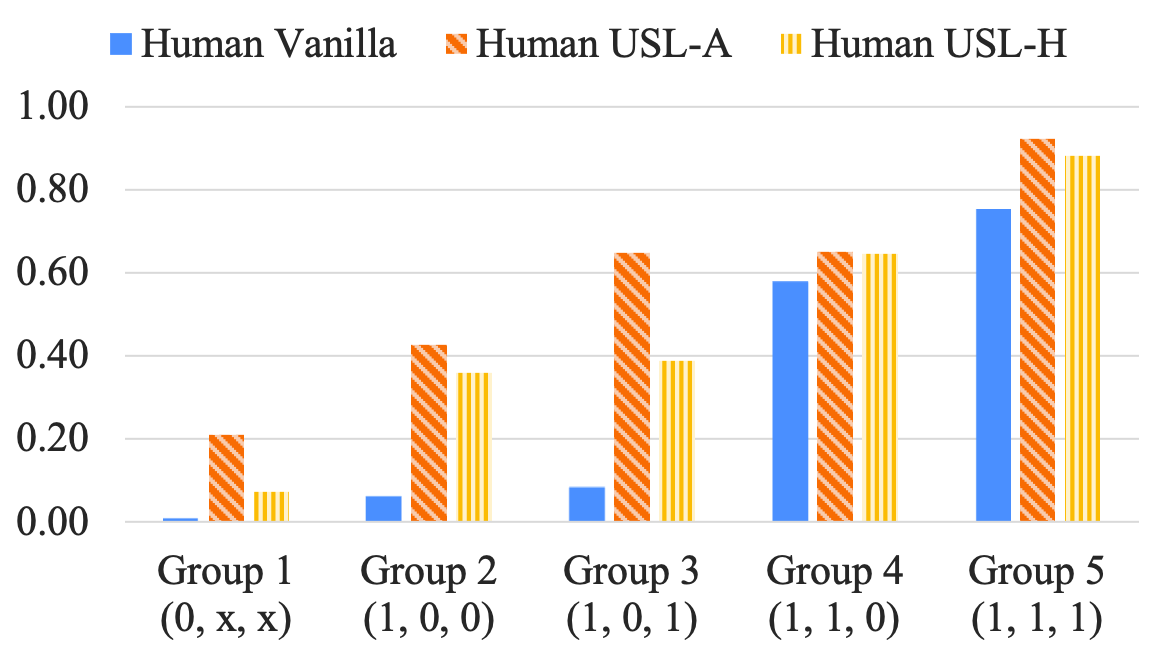}
\centering
\caption{Average of human overall score of the five groups. ($s_U$,$s_S$,$s_{L_s}$) denotes human score for understandability, sensibleness, and specificity, respectively. x denotes any score.}
\label{fig:mean_score}
\end{minipage}
\end{figure}

\subsection{Analysis of Hierarchical Structure}

To understand the relationship between understandability, sensibleness, likability (in this case, specificity), and the overall quality, first we apply linear regression to get the weight of each aspect among each annotator. Then, we apply softmax function on the weights to make them more interpretable. Figure \ref{fig:annot_heatmap} illustrates that sensibleness has the highest weight among all the aspects. This suggests that the annotators tend to rely on sensibleness as a key factor when determining the overall score.

To further investigate how the three aspects affect the overall score, we grouped the responses into five groups in Figure \ref{fig:mean_score}, where each group is denoted by ($s_U$, $s_S$, $s_{L_s}$). Then, for each group, we computed the mean of the annotated overall score (\texttt{Human Vanilla}), and also composited the annotated scores of the three aspects using our hierarchy method (\texttt{Human USL-H}). For comparison, we also used simple averaging (\texttt{Human USL-A}). The result is shown in Figure \ref{fig:mean_score}.

For \texttt{Human Vanilla}, the score for G1 is extremely low compared to the ones of other groups. This means that the score of sensibleness and specificity has no influence when the response is not understandable. Thus, understandability is a crucial building block before other qualities. Also, G2 is almost identical to G3. This suggests that the specificity does not influence the overall score if the response is unsensible. On the contrary, G4 achieves better scores than G3 even though it is entirely unspecific. This validates our hypothesis that sensibleness should be prioritized over specificity in evaluating the overall quality. 

Note that the \texttt{Human Vanilla} scores of G2 and G3 are almost as low as the one of G1, which indicates that even if a response is understandable, it does not significantly affect the overall score unless the response is sensible. We suspect that this problem is due to the subjectivity of annotators because, in a real conversation, it is rare for a speaker to say an incomprehensible utterance. Moreover, we did not provide any concrete instructions on how the overall score should be evaluated. Thus, the annotators fail to consider the understandability aspect properly. Recently, \newcite{mehri2020usr} found a similar result when they ask annotators to evaluate the overall quality with respect to five different aspects. Their study also showed that every annotator prioritizes each quality differently.

Compared to \texttt{Human Vanilla} and \texttt{Human USL-A}, \texttt{Human USL-H} can perform better due to two factors. (1) \texttt{Human USL-H} explicitly considers understandability. It assigns higher scores to G2 as an incentive to a response for being understandable. This makes distinguishing between G1 and G2 easier. (2) It mimics the characteristics of \texttt{Human Vanilla}, especially between G2 and G3, when the unsensible responses deserve the same score. However, this is the opposite with \texttt{Human USL-A} that assigns scores to G3 as high as G4, which contradicts with \texttt{Human Vanilla}. Due to the benefits of \texttt{Human USL-H}, we will use that as the human overall score, unless stated otherwise.

\subsection{Suitable Metrics for Fundamental Aspects}

\begin{table}[t]
\centering
\begin{minipage}[t]{0.48\textwidth}
\centering
\begin{tabular}{llcc} 
    \toprule
    & Metric & Pearson & Spearman \\ \midrule
    \multirow{6}{*}{ \begin{turn}{90}Understandable\end{turn} } 
    & Human (Avg) & 0.4510*   & 0.4510* \\
    & Human (Max) & 0.6969*   & 0.6969* \\ \cmidrule{2-4}
    & Embedding Avg. & 0.1608 & 0.1165 \\ 
    & MLM PPL     & -0.1638*  & 0.0079 \\
    & BertScore & 0.1698* & 0.1590\\
    & BERT-VUP    & \textbf{0.2554*}   & \textbf{0.1370} \\
    \midrule
    
    \multirow{6}{*}{ \begin{turn}{90}Sensibleness\end{turn} } 
    & Human (Avg) & 0.6181* & 0.6181* \\
    & Human (Max) & 0.6826* & 0.6826* \\ \cmidrule{2-4}
    & Vector Extrema & 0.2908* & 0.2973* \\
    & EN-PONE & 0.4917* & 0.4904* \\
    & BERT RUBER & 0.5158* & 0.4964* \\
    & BERT-NUP & \textbf{0.6272*} & \textbf{0.6145*} \\ 
    \midrule
    
    \multirow{6}{*}{ \begin{turn}{90}Specificity\end{turn} } 
    & Human (Avg) & 0.5179* & 0.5179* \\
    & Human (Max) & 0.6430* & 0.6430* \\ \cmidrule{2-4}
    & Embedding Avg. & 0.2196* & 0.2771* \\
    & MLM NCE & -0.2847* & -0.3746* \\
    & MLM SLOR & 0.3488* & 0.4780*\\
    & MLM Likeli. & \textbf{0.4194*} & \textbf{0.4983*} \\
    \bottomrule
\end{tabular}
\caption{\label{table:each_metric_corr} Correlation for each response quality between the human score and automatic evaluation metrics. \textbf{Bold} denotes the best metric for the corresponding quality, and (*) refers to $p<0.01$. }
\end{minipage}\hfill
\begin{minipage}[t]{0.48\textwidth}
\centering
\begin{tabular}{llll} 
    \toprule
    & \multirow{2}{*}{Metric} & \multicolumn{2}{c}{Human Overall} \\
    & & Vanilla & USL\textsubscript{S}-HH \\ \midrule
    \multirow{2}{*}{} 
    & Human (Avg) & 0.7086*   & 0.6205* \\
    & Human (Max) & 0.7720*   & 0.6734* \\
    \midrule
    
    \multirow{5}{*}{ \begin{turn}{90}Word Overlap\end{turn} } 
    & BLEU-2 & 0.0604 & 0.0515\\
    & BLEU-3 & 0.1073 & 0.0951\\
    & BLEU-4 & 0.1073 & 0.0960\\
    & METOER & 0.1795* & 0.1931*\\
    & ROUGE-L & -0.0166 & -0.0410\\
    \midrule
    
    \multirow{4}{*}{ \begin{turn}{90}Embedding\end{turn} } 
    & Embedding Avg. & 0.1886* & 0.2547* \\
    & Greedy Matching & 0.2166* & 0.2386*\\
    & Vector Extrema & 0.3320* & 0.3005*\\
    & BertScore & 0.2102* & 0.2048*\\
    \midrule
    
    \multirow{7}{*}{ \begin{turn}{90}Learning-based\end{turn} } 
    & BERT RUBER & 0.5801* & 0.5693*\\
    & EN-PONE & 0.5545* & 0.5411*\\
    & BERT-NUP & 0.6868* & 0.6701*\\
    & USL\textsubscript{S}-A & 0.6344* & 0.6482*\\
    & + Weighted & 0.5305* & 0.5509*\\
    & USL\textsubscript{S}-H & 0.6847* & 0.6949*\\
    & + Weighted & \textbf{0.7015*} & \textbf{0.6997*}\\
    \bottomrule
\end{tabular}
\caption{\label{table:metric_corr} Pearson correlation between automatic evaluation metrics and two types of human scores on overall quality. Vanilla score refers to a single overall score that the annotators assigned, whereas USL\textsubscript{S}-HH refers to human score obtained using our method. \textbf{Bold} denotes the best metric for each type of overall score, and (*) refers to $p<0.01$. }
\end{minipage}
\end{table}

In this section, we determine which metric is the most suitable for each aspect. We experiment with all the metrics described in Section \ref{sec:models} by comparing their scores with human judgment on understandability, sensibleness, and specificity using Pearson and Spearman rank correlations. Based on Pearson correlations, four highly correlated metrics of each aspect are selected, represented in Table \ref{table:each_metric_corr}. Among the selected metrics, the most suitable ones are BERT-VUP for understandability, BERT-NUP for sensibleness, and MLM-Likelihood for specificity. We notice that MLM-Likelihood and MLM-PPL are not the appropriate measures for understandability. These two metrics tend to assign a high score to repetitive responses (i.e., \emph{I've got a lot of time to get a new place to be a good place to get a new place.}). However, our BERT-VUP metric can recognize and correctly assign a low score to responses with such repetitions. 

BERT-NUP outperforms other metrics in the sensibleness quality. Unlike BERT-RUBER and EN-PONE that obtain embeddings for context and response separately and concatenate them to obtain context-response pair embedding, BERT-NUP combines them into an array of tokens and may utilize the BERT's capability to find contextual patterns between their tokens. 

The MLM-based metrics achieve moderate correlations with the human score on specificity. This may be due to the simple assumption that a response is specific if it contains at least one uncommon word. Furthermore, the language model tends to assign a lower probability to any rare word occurrence, which is consistent with our assumption.

\subsection{Analysis of USL-H Metric}

We select BERT-VUP, BERT-NUP, and MLM-likelihood as the metrics for understandability, sensibleness, and specificity, respectively. Because the MLM-Likelihood score is not between [0,1], we normalize that using MinMax normalization \cite{jain2005score} to ensure consistency between scores. Then, we composite these scores into USL\textsubscript{S}-H score, a variant of USL-H score focusing only on specificity as part of likability. We also implement weighted average ($\alpha_1 s_U + \alpha_2 s_S + \alpha_3 s_L$), similar to \newcite{mehri2020usr}, denoted as USL\textsubscript{S}-A. We utilize the weights obtained from the linear regression (Figure \ref{fig:annot_heatmap}) and assign them as coefficients to $\alpha_1$, $\alpha_2$, and $\alpha_3$. Table \ref{table:metric_corr} shows Pearson correlations between the automatic evaluation metrics with two types of human overall score (vanilla and USL\textsubscript{S}-H). To avoid ambiguity between USL\textsubscript{S}-H score of human and metrics, we denote human USL\textsubscript{S}-H as USL\textsubscript{S}-HH.

Table \ref{table:metric_corr} shows that the weighted USL\textsubscript{S}-H metric outperforms all other baselines; the BERT-NUP metric achieves the second-best performance. This agrees with our hypothesis that incorporating additional information, such as understandability and specificity, with sensibleness score can further enhance the evaluation metric performance. On the other hand, USL\textsubscript{S}-A has lower correlation compared with BERT-NUP and USL\textsubscript{S}-H. This may be because the metric attempts to incorporate the specificity quality, even if the response is incomprehensible or unsensible. This scenario would not occur with our proposed hierarchy since specificity becomes less important as the understandability or sensibleness drops.  

\subsection{Configurability}

\begin{figure}[t]
\centering
\begin{multicols}{3}
    \includegraphics[width=\linewidth]{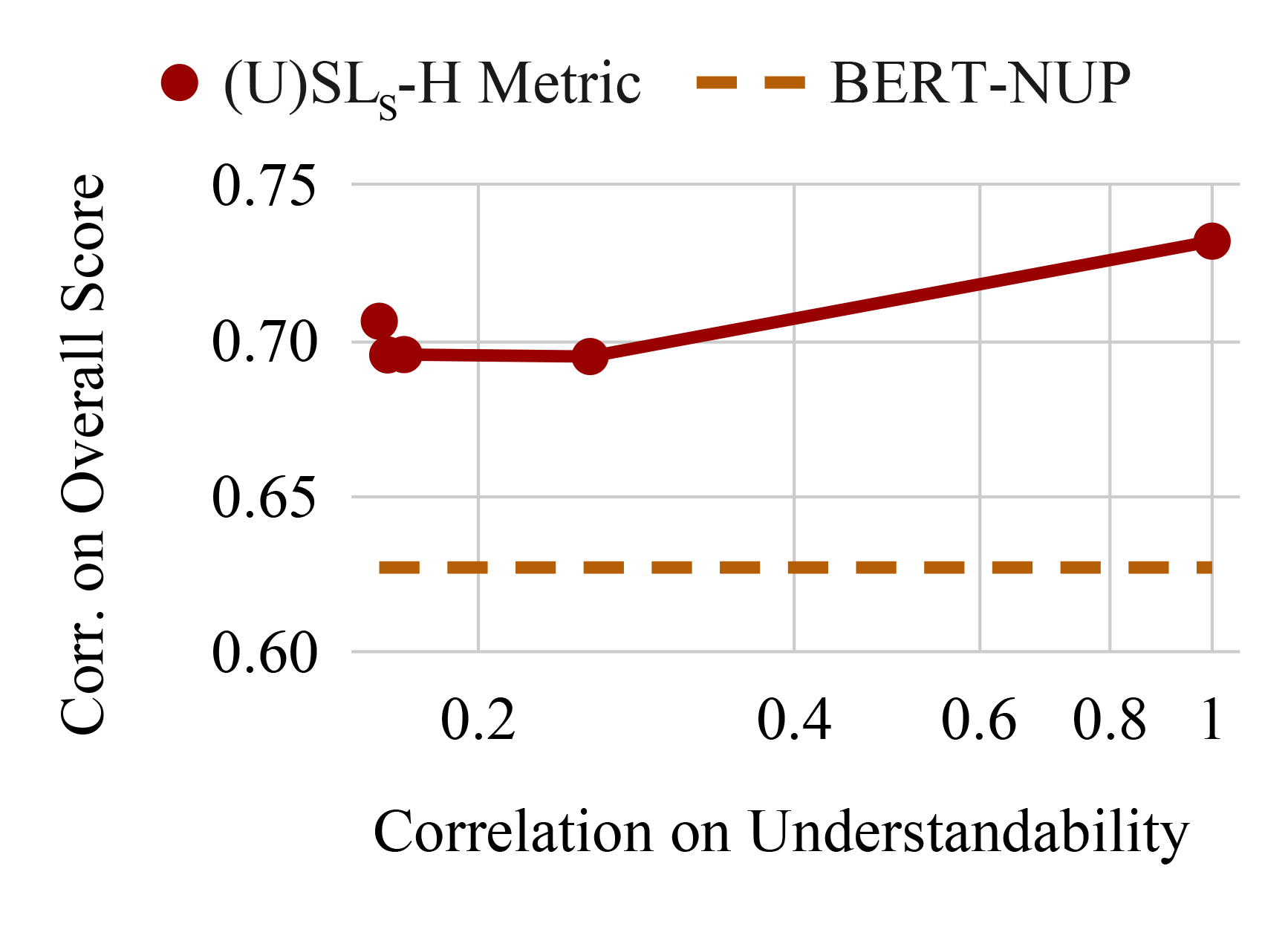}\par
    \includegraphics[width=\linewidth]{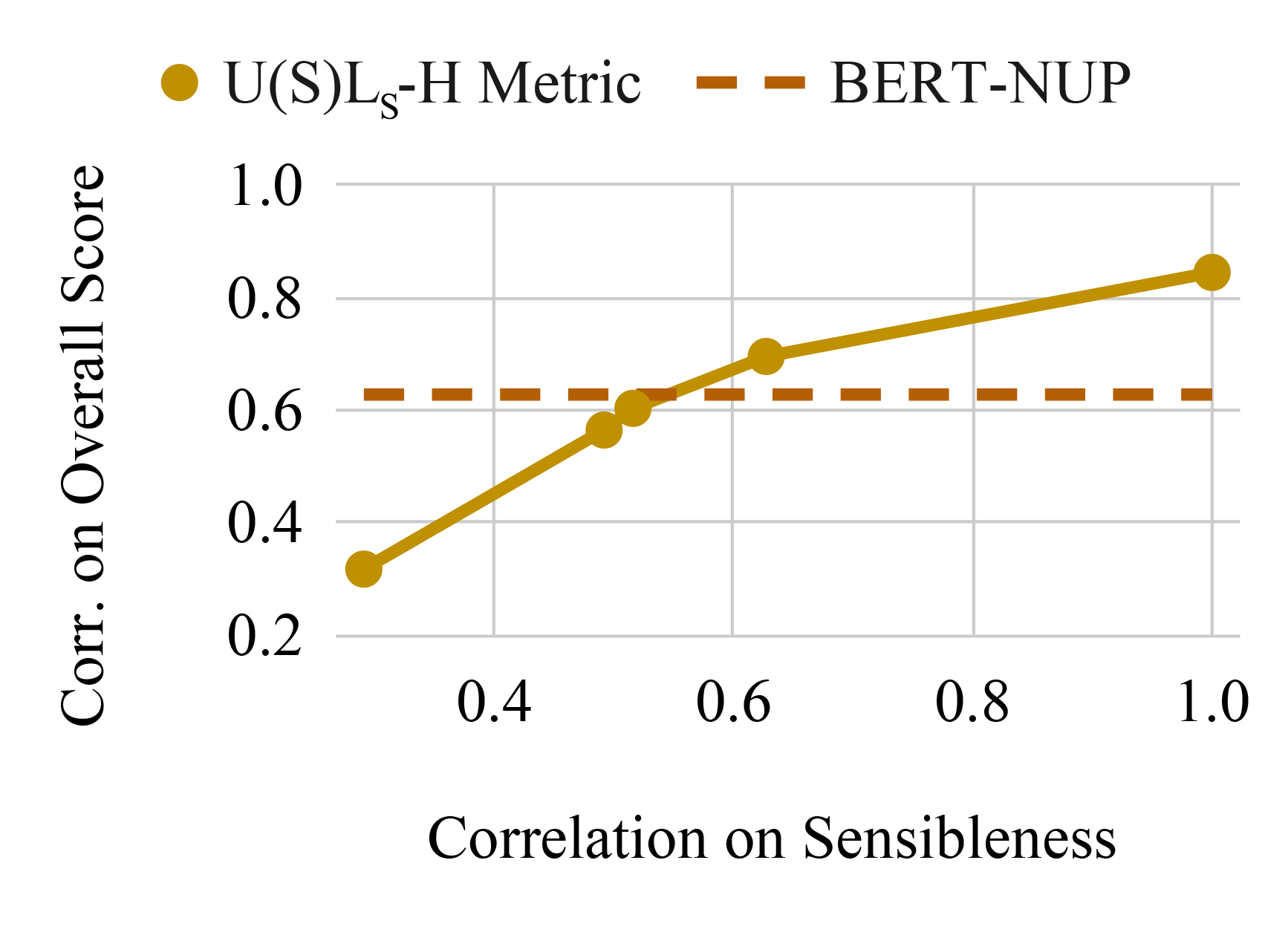}\par
    \includegraphics[width=\linewidth]{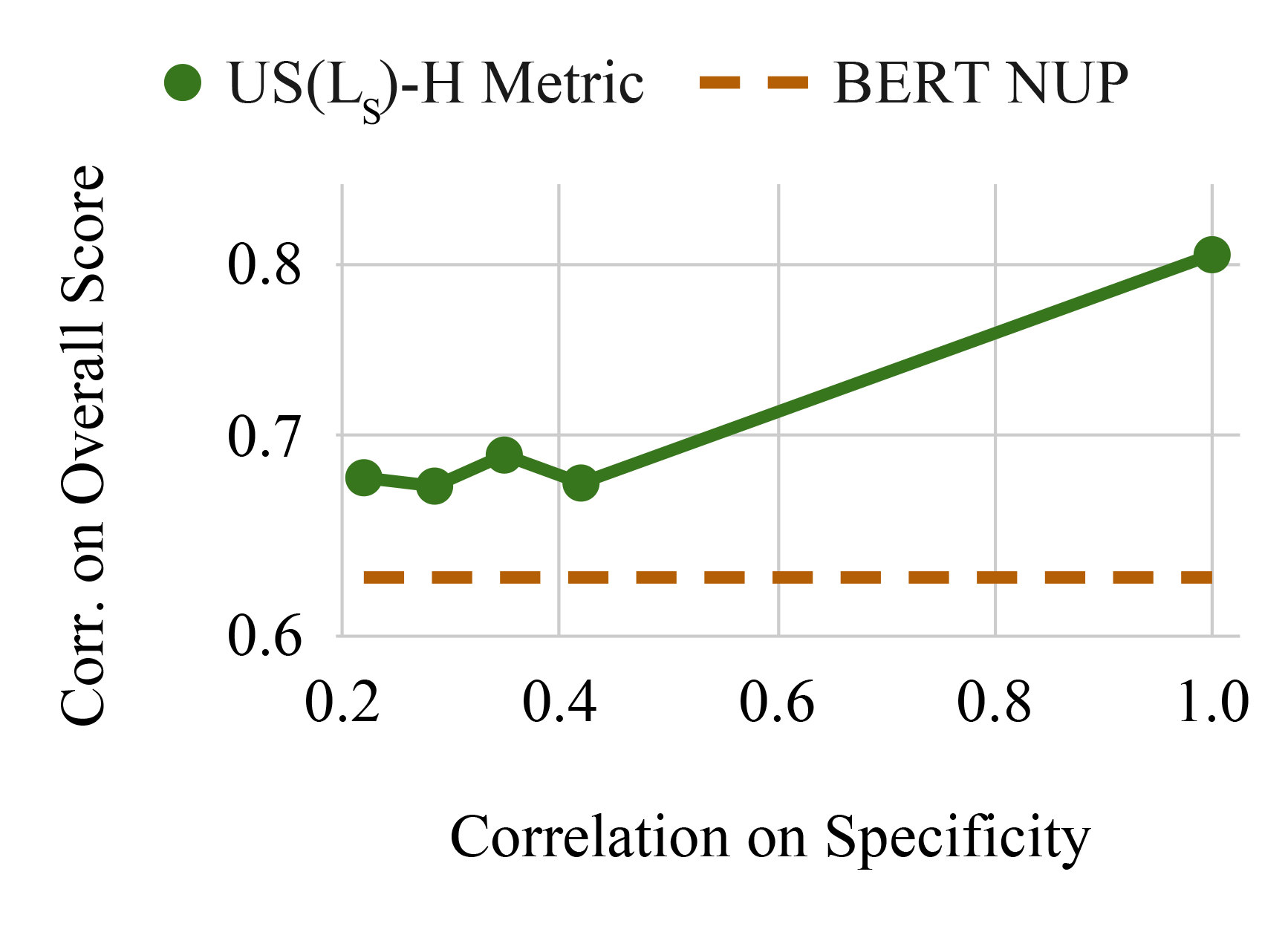}\par
\end{multicols}
\caption{\label{fig:plug-and-play} Pearson correlation between overall score and USL\textsubscript{S}-H metric with changeable sub-metrics for each aspect. Each sub-figure corresponds to different compositions with one sub-metric changeable. The x-axis denotes the correlation of sub-metrics on one aspect. The y-axis denotes the correlation of the USL\textsubscript{S}-H scores. Parentheses (x) denotes that metric of x is changeable. }
\end{figure}

\paragraph{Improving an Aspect} It is uncertain if the USL\textsubscript{S}-H metric can be improved further by utilizing a better sub-metric. Therefore, we tested with a different combination of sub-metrics, each of which has a different correlation. We use BERT-VUP, BERT-NUP, and MLM-Likelihood as the base metrics. To observe the effect of understandability on USL\textsubscript{S}-H, we fix BERT-NUP and MLM-Likelihood constant as we change only the understandability metric. Additionally, we assume that there is an ideal function for each aspect such that they are perfectly correlated with the human score. To obtain such a score, we use the human score itself. We apply this procedure for all three aspects.

Figure \hyperref[fig:plug-and-play]{\ref{fig:plug-and-play}-a}, \hyperref[fig:plug-and-play]{\ref{fig:plug-and-play}-b}, \hyperref[fig:plug-and-play]{\ref{fig:plug-and-play}-c} shows the correlation of metric USL\textsubscript{S}-H with human USL\textsubscript{S}-H, as we change only the understandability, sensibleness, and specificity metric, respectively. Different metrics on Figure \hyperref[fig:plug-and-play]{\ref{fig:plug-and-play}-a} and Figure \hyperref[fig:plug-and-play]{\ref{fig:plug-and-play}-c} do not have any significant impacts on the correlation of the USL\textsubscript{S}-H scores, whereas using a perfectly correlated score does. This does not suggest that these two aspects are insignificant since the performance would decrease drastically if we use only BERT-NUP. Instead, it suggests that the metrics for these aspects may require further improvement to increase the performance of USL\textsubscript{S}-H. Figure \hyperref[fig:plug-and-play]{\ref{fig:plug-and-play}-b}, on the other hand, indicates that the better the sensibleness metric is, the more correlated USL\textsubscript{S}-H will be. Thus, little improvement on sensibleness could also enhance the USL\textsubscript{S}-H.

\paragraph{Swapping an Aspect} To verify whether the USL-H metric is configurable to different aspects, we swap specificity with empathy quality. Then, we trained a BERT-based binary classifier similar to BERT-VUP and grouped the seven emotion labels provided in DailyDialog dataset into two labels: \emph{has emotion} label and \emph{has no emotion} label. We consolidated BERT-VUP, BERT-NUP, and this metric to get another variant of USL-H and denoted it as \emph{USL\textsubscript{E}-H}, whose E stands for empathy. To demonstrate that \emph{USL\textsubscript{E}-H} metric can recognize a sensible and empathetic response better than the other metrics, we use DialoGPT model to generate a pool of 100 responses given a context using two variants of the temperature. We use five metrics to evaluate them. The best response for each metric is selected and is paired against another response selected by another metric to determine which metric selects a better one given the same context. We apply this procedure to 50 different contexts. For each sample, we ask three crowdsource workers to choose \emph{a response that makes more sense and expresses more understanding of the feeling}.  

As shown in Table \ref{table:win_lose}, the human evaluators agree that the responses selected by \emph{USL\textsubscript{E}-H} have higher qualities in terms of sensibleness and empathy, compared to the ones selected by the other metrics. Furthermore, \emph{USL\textsubscript{E}-H} outperforms \emph{USL\textsubscript{S}-H} by a huge margin. This suggests that although \emph{USL\textsubscript{S}-H} achieves good performance with specificity, it does not consider empathy quality. However, we can configure the metric by replacing specificity with the empathy sub-metric to obtain a more suitable variant for the task.

\begin{table}[t]
    \centering
    \begin{tabular}{lcccc}
        \toprule
        Comparison (A vs. B) & Win Rate & A Win & B Win & Tie \\\midrule
        USL\textsubscript{E}-H vs. BERTScore & \textbf{0.67} & 24 & 12 & 14 \\
        USL\textsubscript{E}-H vs. BERT-RUBER & \textbf{0.53} & 19 & 17 & 14 \\
        USL\textsubscript{E}-H vs. BERT-NUP & \textbf{0.56} & 22 & 17 & 11  \\
        USL\textsubscript{E}-H vs. USL\textsubscript{S}-H & \textbf{0.88} & 30 & 4 & 16 \\
        \bottomrule
    \end{tabular}
    \caption{A/B testing by human comparing ``sensible and feeling-expressive'' response pairs that are selected by each metric, reporting wins rate for A over B (excluding ties).}
    \label{table:win_lose}
\end{table}

\section{Conclusion and Future Work}
This study demonstrated a bottom-up approach to building an automatic evaluation metric by deconstructing the overall quality of a response into three fundamental aspects (understandability, sensibleness, and likability), exploring a suitable metric for each aspect, and reconstructing them back to obtain a single metric. However, we restricted the likability aspect to only specificity or empathy. For our future work, we intend to investigate other likability scores, such as engagement or diversity, to ensure that this metric is usable across different tasks and datasets.

\section*{Acknowledgements}
We would like to thank An Tuan Dao, Takuma Udagawa, Thanakrit Julavanich, and Xanh Thi Ho for valuable discussions and the anonymous reviewers for insightful comments. This work was supported by National Institute of Informatics.


\bibliographystyle{coling}
\bibliography{coling2020}
\newpage

\titleformat{\section}{\large\bfseries}{\appendixname~\thesection .}{0.5em}{}
\appendix
\addcontentsline{toc}{section}{Appendices}




\section{Dataset Collection}
\subsection{Building Response Candidates}
\paragraph{Retrieval Methods} The response from this category is expected to be understandable, but may not be relevant to the context. We use TF-IDF and Dual Encoder \cite{lowe2017training} models, using ParlAI \footnote{\url{https://parl.ai/}} for this experiment. During training, the models are provided with 2 candidates (1 correct response and 1 randomly sampled from the training set) and are trained to select the best one. During inference, we follow the method in \cite{liu2016not} and use the whole corpus as candidates, with the correct response removed. 

\paragraph{Generative Methods} Generative models generally can typically produce a response that is somewhat relevant to the given context; however, they sometimes lack in particular qualities, such as specificity, which results in generic and dull responses, e.g., ``I don't know'' or ``Thank you'' \cite{sordoni2015neural,vinyals2015neural,serban2016building}. Therefore, we select a simple seq2seq model with an attention mechanism \cite{Bahdanau2015NeuralMT}, which is also trained using ParlAI. We also use the pretrained DialoGPT \cite{zhang2019dialogpt} model because it is claimed to generate responses that are relevant and context-consistent. 

\paragraph{Human-Generated Response} Using golden data as a response can introduce a bias in the results because the annotator knows the whole context during annotation. Moreover, within the experiment, the number of contexts visible to the models is limited to only a single turn. Hence, we conduct this data collection to ensure fairness. To complete this task, we use Mechanical Turk \footnote{\url{https://www.mturk.com/}} and ask participants to write a response for a given context. To ensure the quality of the responses, we instruct them with the following requirements: (i) the response must have at least 5 words, (ii) the response must not contain any offensive language, and (iii) the response must not contain any emojis. The rejection of response with emoji is because the DailyDialog dataset does not contain them. We want to ensure that the human-generated response remains as close as to the original distribution as much as possible. Subsequently, we use PyEnchant \footnote{\url{http://pyenchant.github.io/pyenchant/}} to detect irregular words and manually correct them.

\subsection{Human Judgement Score}
Before running the actual annotation on collecting human judgements from our volunteers, we conduct two trial runs to verify that our volunteers understand the task and to ensure that the inter-annotator agreement is acceptable.

\section{Further Analysis}

\subsection{Composite Functions} 
Although the human USL\textsubscript{S}-H score explicitly includes understandability, it does not guarantee that the score of this composite function is a suitable replacement for the human overall quality. To ensure that the human USL\textsubscript{S}-H score maintains the quality of the human overall score, we computed their correlation. We also experimented with other composite functions, such as (i) arithmetic mean, (ii) geometric mean, and (iii) harmonic mean, and present the results in Table \ref{table:score_corr}. Using the geometric mean or harmonic mean yields a better correlation than using the arithmetic mean; however, our USL\textsubscript{S}-H score outperforms all these functions on both the Pearson correlation and Spearman rank correlations. This implies that the USL\textsubscript{S}-H score can merge understandability, sensibleness, and specificity explicitly into a single score to reflect the required qualities of a response. 

\begin{table*}[ht]
\centering
\begin{tabular}{c|cccc}
\toprule Correlation & Arithmetic & Geometric & Harmonic & Ours\\
\midrule
Pearson & 0.8038 & 0.8909 & 0.8909 & \textbf{0.9490} \\
Spearman & 0.8059 & 0.8666 & 0.8365 & \textbf{0.9399} \\
\bottomrule
\end{tabular}
\caption{\label{table:score_corr} Correlation between the overall quality and different composite functions on combining human score of each aspect. Every correlation has $p<0.001$.}
\end{table*}

\subsection{Additional Sub-Metric Analysis}
\paragraph{Languge Modeling} We also implemented a 2-layers LSTM-based Seq2Seq model, and it achieves similar performance compared to BERT-based MLM. We choose this BERT-MLM metric over Seq2Seq model for two reasons: (i) BERT can generalize well to other datasets, and (ii) to maintain consistency with other BERT-VUP and BERT-NUP metrics in our work.

\paragraph{Inverse Word Frequency} We also experiment with the inverse word frequency metric proposed by \cite{zhang2018learning}. However, based on Pearson correlation score, this metric was not in the top four.

\section{Case Studies} 
Table \ref{table:example} consists of some examples to compare the USL\textsubscript{S}-H score with other baselines. The METEOR metric does not perform well due to few words overlapped between golden and candidate responses. Not only are the two responses lexically different, but also they are semantically different. This yields a relatively low score for BERTScore. On the other hand, BERT-RUBER metric assigns a better score compared with the former metrics; however, it does not consider the understandability aspect. For example, BERT-RUBER metric assigned a 0 score to example 3 when the response is understandable and deserved a score higher than 0. The USL\textsubscript{S}-H metric is on par with the human score. It recognizes when the response is incomprehensible, unsensible, or unspecific and can present these qualities in an interpretable manner. However, combining three sub-metrics into a single metric has one limitation. It requires every sub-metric to perform well; otherwise, it will not attain a score of 1.00, as shown in example 1. 

\begin{table}[ht]
\small
\begin{tabularx}{\textwidth}{p{0.16\textwidth}|p{0.14\textwidth}|p{0.17\textwidth}|Y|Y|Y|Y|Y}
\toprule
  Context & Ground-Truth Response & Candidate Response & Meteor & BERT
  Score& RUBER & \text{USL\textsubscript{S}-H} & Human\\
\midrule


Can you tell me how to get to the Jade Buddha Temple? &
I'm sorry. I can't quite follow you. Would you mind speaking slowly? &
Go straight ahead until you see the roundabout, then take a left turn. &
0.03 & -0.16 & 1.00 & 0.78 & 1.00 \\ \midrule

Let' s play a game! &
OK! How about Scrabble? &
Sure. &
0.00 & 0.08 & 0.41 & 0.55 & 0.66 \\ \midrule

Sir what's the boarding time? &
I am sorry. The flight is late... 
&
Are you trying to act like a real man?  &
0.02 & -0.04 & 0.00 & 0.33 & 0.33 \\ \midrule

Excuse me, sir, but are you Mr.Richard of World Trading Company? &
Yes. That' s right. &
I'm afraid I'm afraid I'm afraid I'm afraid I'm afraid I can't. &
0.05 & -0.20 & 0.41 & 0.04 & 0.08 \\
\bottomrule
\end{tabularx}
\caption{\label{table:example} Examples of different metrics. BERTScore is within the range [-1, 1], whereas the range of the other metrics is in [0,1]. The human score in this case refers to the human score of USL\textsubscript{S}-H.}
\end{table}


\end{document}